\documentclass[twoside,11pt]{article}

\usepackage{jmlr2e}

\usepackage{caption}
\usepackage{subcaption}
\usepackage{tablefootnote}
\usepackage{graphicx}
\usepackage[outdir=./Figures/]{epstopdf}
\usepackage{algorithm}
\usepackage[noend]{algpseudocode}

\usepackage{multirow}
\usepackage{booktabs}

\usepackage{amsmath}
\usepackage{amssymb}
\usepackage[T1]{fontenc}
\usepackage{graphicx}
\allowdisplaybreaks

\usepackage{enumitem}
\setitemize{noitemsep,topsep=0pt,parsep=0pt,partopsep=0pt}

\DeclareMathOperator*{\argmax}{arg\,max}

\ShortHeadings{A Contextual-bandit-based Approach for Informed Decision-making in Clinical Trials}{ }
\firstpageno{1}

\begin{document}

\title{A Contextual-bandit-based Approach for Informed Decision-making in Clinical Trials}

\author{\name Yogatheesan Varatharajah \email varatha2@illinois.edu \\
	\addr Electrical and Computer Engineering, University of Illinois, Urbana, Illinois 61801. 
	\AND
	\name Brent Berry \email Berry.Brent@mayo.edu \\
	\addr Department of Neurology, Mayo Clinic, Rochester, Minnesota 55904.
	\AND
	\name Sanmi Koyejo \email sanmi@illinois.edu \\
	\addr Computer Science, University of Illinois, Urbana, Illinois 61801.
	\AND
	\name Ravishankar Iyer \email rkiyer@illinois.edu \\
	\addr Electrical and Computer Engineering, University of Illinois, Urbana, Illinois 61801.} 

\maketitle

\vspace{0.2in}

\begin{abstract}
	Clinical trials are conducted to evaluate the efficacy of new treatments. Clinical trials involving multiple treatments utilize randomization of the treatment assignments to enable the evaluation of treatment efficacies in an unbiased manner. Such evaluation is performed in post hoc studies that usually use supervised-learning methods that rely on large amounts of data collected in a randomized fashion. That approach often proves to be suboptimal in that some participants may suffer and even die as a result of having not received the most appropriate treatments during the trial. Reinforcement-learning methods improve the situation by making it possible to learn the treatment efficacies dynamically during the course of the trial, and to adapt treatment assignments accordingly. Recent efforts using \textit{multi-arm bandits}, a type of reinforcement-learning methods, have focused on maximizing clinical outcomes for a population that was assumed to be homogeneous. However, those approaches have failed to account for the variability among participants that is becoming increasingly evident as a result of recent clinical-trial-based studies. We present a \textit{contextual-bandit}-based online treatment optimization algorithm that, in choosing treatments for new participants in the study, takes into account not only the maximization of the clinical outcomes but also the patient characteristics. We evaluated our algorithm using a real clinical trial dataset from the International Stroke Trial. The results of our retrospective analysis indicate that the proposed approach performs significantly better than either a random assignment of treatments (the current gold standard) or a \textit{multi-arm-bandit}-based approach, providing substantial gains in the percentage of participants who are assigned the most suitable treatments. The contextual-bandit and multi-arm bandit approaches provide 72.63\% and 64.34\% gains, respectively, compared to a random assignment.


\end{abstract}

\section{Introduction}
\label{sec:intro}
\indent A randomized clinical trial is the current gold-standard approach for evaluating treatment efficacy. In such a setting, the participants are divided randomly into separate groups to enable comparison of different treatments or other interventions. Since these trials usually require large sample sizes and therefore long study durations, a large number of participants receive suboptimal treatments, especially when multiple treatments are being evaluated in a trial \citep{fuster2012guided, cummings2014alzheimer, minnerup2014analysis}. Although some information about the efficacies of treatments and their relation to patient characteristics is acquired during the course of these trials, this information is almost always underutilized. That is a big limitation because, if correctly utilized, this information has the potential to provide huge monetary savings and improved patient outcomes \citep{berry1995adaptive}. This paper introduces a decision-theoretic approach to address that limitation and shows its utility in a real clinical trial dataset.

Several recent studies have proposed adaptive strategies for treatment administration in clinical trials \citep{villar2015multi, villar2018covariate}. These approaches achieve adaptability by utilizing a type of online reinforcement learning algorithms known as \textit{multi-arm bandits} \citep{gittins2011multi}. Initial developments on multi-arm bandits occurred in the context of gambling scenarios in which an agent had to choose an action that would maximize the rewards, and the strategy (or policy) for choosing the actions was dynamically updated during the game. That problem involved exploration/exploitation tradeoffs: there should be a balance between trying different actions to learn more about their expected payouts, and wanting to exploit the best action based on the information already obtained. \textit{Bandit} approaches employ different sampling strategies to effectively handle this tradeoff. In the setting of a clinical trial, such an algorithm aims to identify a policy for assigning patients into treatment subgroups in a way that maximizes favorable clinical outcomes while still allowing the exploration of previously under-explored treatments. 

Although the general approach of multi-arm bandits perfectly suits our goal, a limitation of the prior studies utilizing this approach is that they do not account for the inter-patient variability, e.g., how two patients may respond to the same treatment differently. Instead, they consider all patients to be similar in the way they respond to treatments and they therefore come up with a common dynamic policy for all new participants. There is increasing evidence in the medical literature that patient populations are heterogeneous and that individuals respond to treatments differently \citep{athreya2017data}. This phenomenon especially pronounced even more profound in complex neurological and cardiovascular diseases for which the same clinical diagnosis might arise from different underlying pathological mechanisms \citep{molina2008clinical}. Hence, an online learning algorithm that considered disease-related characteristics of the participants in addition to the general multi-arm bandit setting would be better suited for clinical trials. In this paper, we describe a \textit{contextual-bandit}-based approach that incorporates patient characteristics to refine treatment select in clinical trials. 

\textit{Contextual bandits} \citep{li2010contextual} are a generalization of multi-arm bandits in which the policy for choosing future actions is dependent on the context of the game. In the setting of a clinical trial, the context can be the disease-related characteristics of a new participant that determine which of the treatments will be beneficial for that participant. From a different perspective, the contextual-bandit approach attempts to dynamically stratify participants based on their predispositions to treatment responses. We showcase the utility of this model using a publicly available dataset collected during the International Stroke Trial (IST), a clinical trial that evaluating the efficacies of two drug-based treatments in altering the course of acute ischemic stroke \citep{carolei1997international}. We imitated the real-time scenario by sequentially going through the data of each participant in the trial. Based on the treatment given to each participant and the corresponding clinical outcome, we utilized our approach to learn about and update our understanding of the relationship between contexts (i.e., patient characteristics), treatments, and clinical outcomes. Then we used that knowledge to choose treatments to administer to the next participant. We repeated that process until the end of the trial and recorded the clinical outcomes resulted from our approach. The results of this retrospective analysis indicate that the contextual-bandit-based approach performs significantly better than a either random assignment of treatments (the current gold standard) and a context-free multi-arm-bandit-based approach, providing substantial gains in the percentage of participants who received the most suitable treatments. The contextual-bandit and multi-arm bandit approaches provided 72.63\% and 64.34\% better results, respectively, than the random assignment did.

\section{Related work}
\label{sec:rel_work}

There has been significant interest in developing adaptive strategies for clinical trials \citep{coffey2008adaptive, berry2010adaptive}. Adaptive trials allow specific changes in key trial attributes (e.g., sample size, the test statistic, or the outcome variable used to measure the treatment effect) across the course of the trial based on information acquired during the trial \citep{pocock1977group}. However, most adaptive trials proposed in the past focused on evaluating a single treatment in a single population \citep{freidlin2005adaptive}. In addition, traditional designs of adaptive strategies have primarily concentrated on the statistical attributes of the trials and have neglected to consider the well-being of participants or the cost of ineffective treatments. 

Adaptive strategies that evaluate multiple treatments and combinations of treatments have received interest recently \citep{villar2015multi, villar2018covariate}. Although they take into account the well-being of participants and the cost of ineffective treatments, these multi-arm-bandit-based approaches do not generalize to heterogeneous patient populations in which participants with similar medical conditions might respond to treatments differently. Many recent posthoc studies of clinical trials have demonstrated that treatment responses of individuals with similar clinical conditions vary based on their clinical and biomarker profiles \citep{catenacci2015next, lazar2016identifying, marshall2014have}. Such recent revelations necessitate context-aware strategies in adaptive clinical trials. A contextual-bandit is a light-weight reinforcement learning approach that is suitable for learning behavior from feedback when the behavior depends on external factors in addition to the stimuli \citep{littman2015reinforcement}. Contextual-bandit approaches have been successfully applied in recommender systems, where the recommendations to a specific user depend on the feedback given by the user for the past recommendations as well as user-specific characteristics \citep{li2010contextual, li2011scene}. Our approach, to our knowledge, is the first to apply it on a clinical trial setting and demonstrate its efficacy using a real clinical trial dataset.

\section{Model description}
\label{sec:mod_desc}
\begin{figure*}[h]
	\centering
	\includegraphics[width=0.7\linewidth]{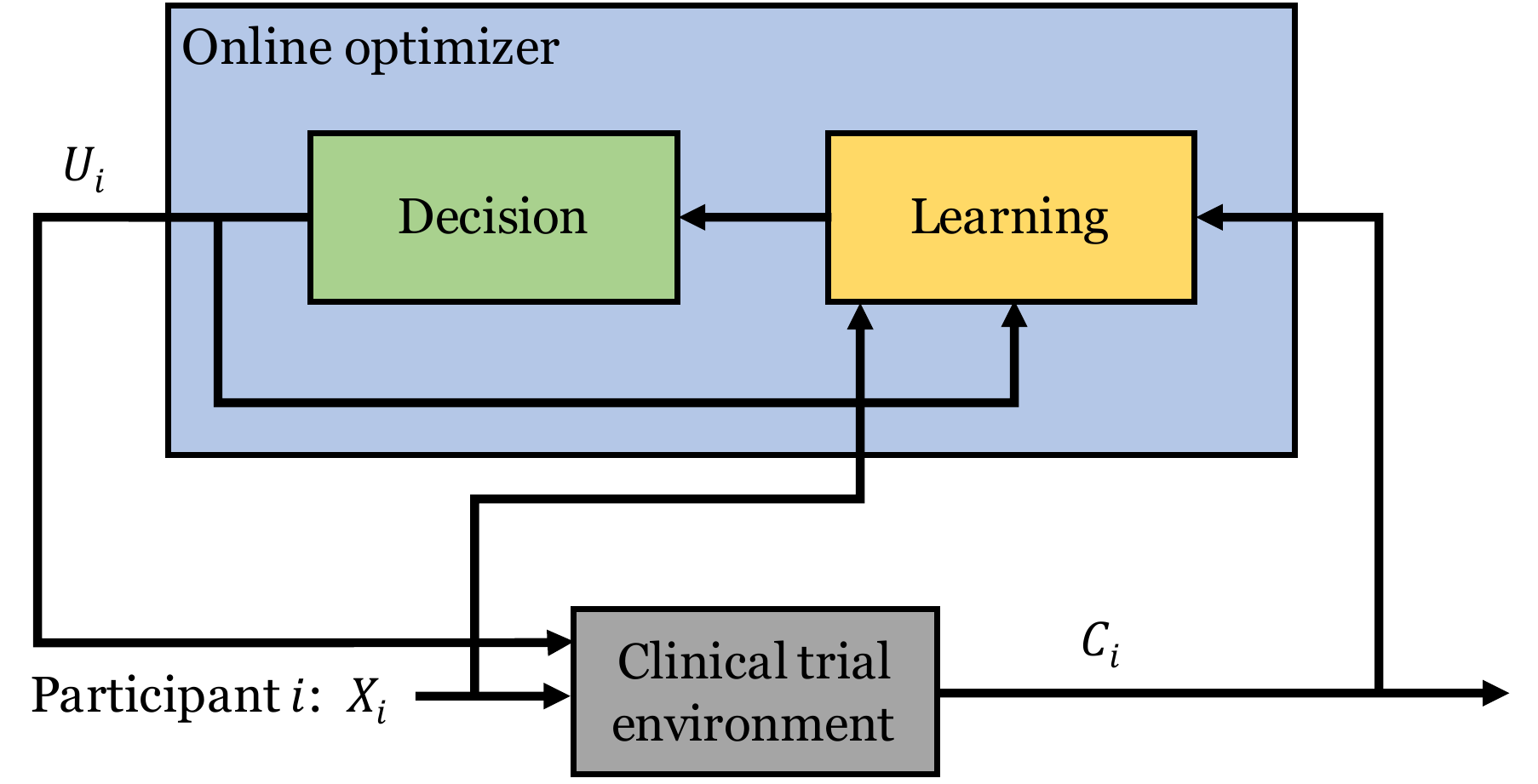}
	\caption{Illustration of the online optimizer used in a clinical trial setting. }
	\label{fig:online_learner}
\end{figure*}

\noindent \textbf{Definitions:} Consider a clinical trial involving $K$ treatments.The context of patient $i$ is denoted by $X_i \in \mathbb{R}^D$, which is a vector consisting of $D$ attributes related to the disease being considered in the trial. We also use $U_i$ to denote the treatment that was provided to patient $i$, and $C_i$ to denote the corresponding clinical outcome, where $U_i$ is a discrete random variable taking values in the set $\{1,\ldots,K\}$. In addition, we assume that the clinical outcomes ($C_i$) can be dichotomized as successes or failures and model them as binary random variables, i.e., $C_i \in \{0,1\}$. We are interested in choosing a treatment for each participant of the trial based on his or her specific disease-related characteristics in order to maximize the probability of successful recovery. We do so by using an online optimizer that involves a learning component to learn from past information and a decision component to choose optimal actions based on learned response patterns. That process is illustrated in figure \ref{fig:online_learner}. 

\noindent \textbf{Context-free setting:} Here we will first formalize the problem for a setting that does not take the context into account and then we will extend it to a contextual setting later in this section. In the context-free setting, the outcomes depend only on the treatment provided. Therefore, the outcomes can be seen as having been drawn from a Bernoulli distribution with parameter $\theta_u$, where $\theta_u$ denotes the success probability (or the mean reward) of treatment $U=u$. The mean rewards $\theta = \{\theta_1,\ldots, \theta_K\}$ are unknown, but not time-varying.

\noindent \textbf{Objective:} Given the above definitions, our goal is to choose the treatments in such a way that they are optimal for each patient. Suppose that the optimal treatment for patient $i$ is $U_i^\star$ and the corresponding outcome is $C_i^\star$. Then, the objective of the online optimizer is to minimize the following two quantities, where $N$ denotes the total number of participants in the study.
\begin{align}
\label{eq:obj_fun}
R &= \sum_{i=1}^{N} \left[C_i^\star-C_i\right] \\
S &= \sum_{i=1}^{N} \left[\mathbb{I}_{U_i^\star\neq U_i}\right]
\end{align} 
where $\mathbb{I}$ is the indicator function. 

The first quantity $R$ (also known as \textit{regret}) measures the difference between the actual outcomes from the optimal outcomes that would have been achieved had we identified the correct treatment for each patient. The second quantity $S$ (also known as the \textit{suboptimal action count}) measures the number of instances when the chosen treatment was not the same as the correct treatment for a patient. 

\noindent \textbf{Model of the priors:} In order to achieve the above goal, we are interested in learning the correct priors for each treatment $u$, and utilizing those priors to choose treatments for the new participants in the study. We model the priors to be beta-distributed with parameters $\alpha=\{\alpha_1,\ldots,\alpha_K\}$ and $\beta=\{\beta_1,\ldots,\beta_K\}$. Please note that the beta distribution is the conjugate prior of the Bernoulli distribution. Then, for each treatment $u$, the probability density function of $\theta_u$ is
\begin{equation}
p(\theta_u) = \frac{\Gamma(\alpha_u+\beta_u)}{\Gamma(\alpha_u)\Gamma(\beta_u)}\theta_u^{\alpha_u-1}(1-\theta_u)^{\beta_u-1},
\end{equation}
where $\Gamma$ denotes the Gamma distribution. We begin with an independent prior belief over each $\theta_u$, i.e., $\alpha_u=\beta_u=1$ for each $u\in\{1,\ldots,K\}$. With each new observation, (i.e., a treatment ($U_i$), outcome ($C_i$) pair), the distribution is updated using Bayes' rule. Because of the conjugacy properties, the parameters $\alpha_u$ and $\beta_u$ can be updated using the following simple rule.
\begin{equation}
(\alpha_u,\beta_u)\leftarrow \left\{\begin{matrix}
(\alpha_u,\beta_u)& \text{if } U_i\neq u\\ 
(\alpha_u,\beta_u)) + (C_i,1-C_i)&  \text{if } U_i=u
\end{matrix}\right.
\end{equation}
Note that a beta distribution with parameters $(\alpha_u,\beta_u)$ has the mean $\alpha_u/(\alpha_u+\beta_u)$, and that the distribution becomes more concentrated as $\alpha_u$ and $\beta_u$ become large. In general, this formulation is known as the \textit{Bernoulli bandit}.

\noindent \textbf{Choosing the treatment for a new patient:} Since $\theta_u$s are beta-distributed, a naive choice of treatment for a new patient is the treatment whose prior has the largest mean, i.e., $U_i = \argmax_u \left[\frac{\alpha_u}{\alpha_u+\beta_u}\right]$. Although that greedy approach is a valid choice, a downside is its inability to balance the exploration/exploitation trade-off \citep{russo2017tutorial}. Here we describe two popular bandit algorithms for choosing the treatment for a new patient based on the distributions of priors learned from past experiments. Both are extremely effective in balancing exploration/exploitation trade-off. 

\noindent \textbf{Thompson sampling} is a Bayesian approach \citep{agrawal2012analysis} that randomly samples the success probabilities of treatments from their respective prior distributions and selects a treatment with the maximum sample value. It is easy to see that the distributions of the priors will be more spread at the beginning of the trial, and therefore that all the treatments will be selected with similar probabilities. With more participants, the distributions will become narrower and the treatments with higher rewards are more likely to be selected. However, unlike the greedy case, treatments with low rewards will still be selected with relatively low probabilities, because the approach uses samples drawn from the distributions of success probabilities. In that way, the Thompson sampling algorithm effectively balances the exploration/exploitation trade-off. This is illustrated for the Bernoulli bandit case in Algorithm \ref{thompson}.

\noindent \textbf{The Upper confidence bound (UCB)} algorithm, on the other hand, is a frequentist approach that uses point estimates of the success probabilities of treatments to choose future treatments. It also uses an extra additive term ($\sqrt{\frac{\text{ln}i}{n_{u,i}}}$, which is added to the point estimates of the priors; see Algorithm \ref{ucb}) that is inversely proportional to the number of times a particular treatment has been applied. This additive term is also a function of the duration of the trial and establishes an upper confidence bound for the point estimate \citep{auer2002using}. It starts by applying each treatment at least once and then chooses future actions based on the upper confidence bounds of the treatment success probabilities. As with the Thompson sampling approach, the distributions of the priors will be more spread at the beginning and will become progressively narrower. However, the UCB algorithm handles the exploration/exploitation trade-off slightly differently. The confidence bound of the treatments that have been previously under-explored will grow with the duration of the trial and will eventually exceed the bounds of other treatments and therefore get a chance to be applied to a new patient. However, the chance that the treatments with low rewards will be applied will diminish and eventually vanish. This is illustrated for the Bernoulli bandit case in Algorithm \ref{ucb}.

\begin{algorithm}[H]
	\caption{Thompson sampling}\label{thompson}
	\begin{algorithmic}[1]
		\For{$i=1,2,\ldots,$}					
		\For{$u \in \{1,\ldots,K\}$}
		\State Sample $\hat{\theta_u} \sim beta(\alpha_u,\beta_u)$
		\Comment{sample model}
		\EndFor
		$U_i = \argmax_u \hat{\theta_u}$
		\Comment{select and apply action}
		\State{Apply $U_i$ and observe $C_i$}
		\State{$(\alpha_{U_i},\beta_{U_i})\leftarrow(\alpha_{U_i},\beta_{U_i}) + (C_i,1-C_i)$}
		\Comment{update distribution}
		\EndFor
	\end{algorithmic}
\end{algorithm}

\begin{algorithm}[H]
	\caption{Upper confidence bound (UCB)}\label{ucb}
	\begin{algorithmic}[1]
		\For{$i=1,2,\ldots,K$}	
		\State{Apply treatment $i$}		
		\Comment{apply each treatment once}		
		\EndFor
		\For{$i=K+1,K+2,\ldots,$}	
		\For{$u \in \{1,\ldots,K\}$}
		\State{Estimate $\hat{\theta_u} = \left[\frac{\alpha_u}{\alpha_u+\beta_u}\right]$}
		\Comment{estimate mean rewards}
		\State{$n_{u,i}\leftarrow$ \# of times treatment $u$ has been applied so far}
		\EndFor		
		\State{$U_i = \argmax_u \left[\hat{\theta_u} + \sqrt{\frac{\text{ln}i}{n_{u,i}}}\right]$}
		\Comment{select and apply action}
		\State{Apply $U_i$ and observe $C_i$}
		\State{$(\alpha_{U_i},\beta_{U_i})\leftarrow(\alpha_{U_i},\beta_{U_i}) + (C_i,1-C_i)$}
		\Comment{update distribution}
		\EndFor
	\end{algorithmic}
\end{algorithm}

\noindent \textbf{Contextual setting:} So far, we have considered a setting in which the context of the participant is ignored. However, many recent clinical trials have shown that treatment responses are very much context-dependent. In this paper, we assume that the contexts $X_i$ are $D$-dimensional binary vectors resulting in $2^D$ different contexts. But fortunately, unlike the success probabilities of treatments, the contexts are observable. The success probabilities of treatments are likely different in different contexts. We incorporate the contexts into our approach by treating each context as a different context-free multi-arm bandit problem. Therefore, when a new participant joins the trial, depending on the observed context, a treatment will be chosen based on the context-free bandit problem that includes only the past participants with that specific context. Alternatively, we can maintain $2^D$ number of distinct bandits (one for each context) and choose treatments based on the bandit corresponding to the observed context. This is illustrated in Algorithm \ref{context_bandit}, in which $MAB(m)$ denotes the context-free multi-arm bandit associated with context $m$.

\begin{algorithm}[H]
	\caption{Contextual bandit for clinical trial optimization}\label{context_bandit}
	\begin{algorithmic}[1]
		\For{$m=1,2,\ldots,2^D$}					
		\Comment{initialize all context-free bandits}
		\State{$MAB(m)\leftarrow$initialize context-free bandit()}		
		\EndFor
		\For{$i=1,2,\ldots,$}					
		\State{$X_i\leftarrow$observe context(patient $i$)}
		\State{$MAB\star=MAB(X_i)$}
		\Comment{bandit associated with context $X_i$}
		\State{$U_i\leftarrow$select treatment($MAB\star$)}
		\Comment{select a treatment based on priors in $MAB\star$}
		\State{Apply $U_i$ and observe $C_i$}
		\State{update prior($U_i$, $MAB\star$)}
		\Comment{update the prior of $U_i$ in $MAB\star$}
		\EndFor		
	\end{algorithmic}
\end{algorithm}


\section{Application of the model to International Stroke Trial (IST) database}
\label{sec:prob_desc}

\noindent \textbf{Data:} The International Stroke Trial (IST) was one of the largest randomized trials ever conducted for acute stroke \citep{carolei1997international}.  The IST dataset includes data on 19,435 patients with acute stroke, with 99\% complete follow-up. For each randomized patient, the variables assessed at randomization, at the early outcome point (14 days after randomization or prior discharge), and at 6 months were collected. The primary outcomes that were recorded in the study are death within 14 days and death or dependency at 6 months. The aim of the trial was to establish whether early administration of aspirin, heparin, both, or neither influenced the clinical course of acute ischemic stroke. 

\noindent \textbf{Background:} Stroke is a major source of economic burden and personal hardship to those afflicted. Each year, approximately 795,000 people in the United States suffer a stroke and about 600,000 of them are the person's first stroke \citep{rosamond2008heart}. Unfortunately, because of changing demographics and the fact that stroke is an age-related disease, the prevalence is expected to increase, given the advancing age of the population. Stroke accounts for 1 of every 19 deaths in the U.S., making it the third leading cause of death (behind heart disease and cancer) and in fact it is the leading cause of long-term disability in the U.S. \citep{rosamond2008heart}.  Stroke imposes a huge burden on the economy. The total direct and indirect costs are more than \$100 billion per year counting hospitalization, transition care and rehabilitation care, physician expenditures, medications, ancillary staff and home care, and therapy, as well as indirect costs such as loss of economic productivity \citep{rosamond2008heart}. 

\noindent \textbf{Ischemic strokes:} Approximately 90\% of all strokes are ischemic \citep{adams1993classification}. Ischemic strokes comprise a variety of conditions in which blood flow to part of the brain is reduced, resulting in tissue damage or death, and is usually an acute process. It is extremely critical to obtain timely medical help for ischemic stroke. Untreated ischemic strokes can lead to fluid buildup, swelling, and bleeding in the brain, seizures, and permanent problems with memory and understanding. In addition, there is a 5--17\% risk that another stroke will follow a transient ischemic stroke within three months \citep{adams2007guidelines}. Furthermore, there is substantial evidence that stroke patients with certain other co-morbidities, such as atrial fibrillation, respond to certain types of treatments better than others \citep{paciaroni2007efficacy}. Therefore, it is necessary to obtain timely medical help and the right kind of treatment to avoid additional complications. Hence, an adaptive clinical trial setting may make it possible to learn these complex relationships and provide patients with the right kinds of treatments.
\section{Experiments \& Evaluation}
\label{sec:experiment}
\noindent \textbf{Contextual bandit setting:} Here we describe the steps we took to apply the contextual bandit model we described in Section \ref{sec:mod_desc} to the clinical trial data from the International Stroke Trial. The trial included drug treatments based on aspirin and heparin. Hence, there were four different possible treatments reflecting the different combinations of the two drugs that could have been administered, i.e., $K=4$. In addition, we use the 2-week mortality of the participants to determine clinical outcomes. We consider discharge of a participant from the hospital alive within two weeks to mean the treatment was successful. As previously explained in Section \ref{sec:prob_desc}, another cardiovascular comorbidity, atrial fibrillation can modulate the response to heparin-based treatments. Therefore, we use the binary variable representing whether or not the participant had atrial fibrillation as the context in our algorithm. 

\noindent \textbf{Analytic scheme:} We simulated the online setting by sequentially going through the dtaa of each participant admitted to the trial. Two bandits (one for each context) were created, with 4 choices of treatments. For a new participant in the trial, depending on the context, one of the bandits was selected. Then, we took three different approaches to choose a treatment: a) a random choice (i.e., the strategy currently used in clinical trial settings), b) a Thompson sampling-based approach, and c) a UCB-based approach. It was difficult to estimate the outcome for a treatment chosen by any of the three approaches when that treatment was not the one actually performed in the real clinical trial for that participant. To circumvent that issue, for each context, we used all the participants in the clinical trial dataset who had the same context to obtain estimates of the success probabilities of each treatment and used a Bernoulli sample generator to generate an outcome for each treatment. Note that the success probabilities of each context were calculated separately by considering only the participants with the same context. Those estimated outcomes were used to update the prior distributions within the bandit corresponding to the context of each participant. We repeated that process through the end of the trial and recorded the outcomes and the chosen treatments for each approach. We also evaluated a context-free multi-arm-bandit-based approach, using the same dataset, to showcase the benefits of our approach. In the context-free case, we calculated the success probabilities for the Bernoulli sampler using the whole clinical trial dataset in a context independent manner.

\noindent\textbf{Evaluation:} We utilized the two quantities defined in Section \ref{sec:mod_desc}, i.e., \textit{regret} and \textit{suboptimal action count}, to evaluate our approaches. Since we do not know what would have been the optimal treatment for each participant, we chose the maximum of all the success probabilities estimated using the whole dataset as the optimal outcome and the corresponding treatment as the optimal treatment. In the context-free case all the participants had the same optimal treatment and corresponding outcome; in the contextual case, the participants in each context had a their own identified optimal treatment and a corresponding outcome. 

\section{Results}
\label{sec:results}
\begin{figure*}[h]
	\centering
	\begin{subfigure}[t]{0.49\textwidth}
		\centering
		\includegraphics[trim=0cm 6cm 0cm 6cm, clip, width=\linewidth]{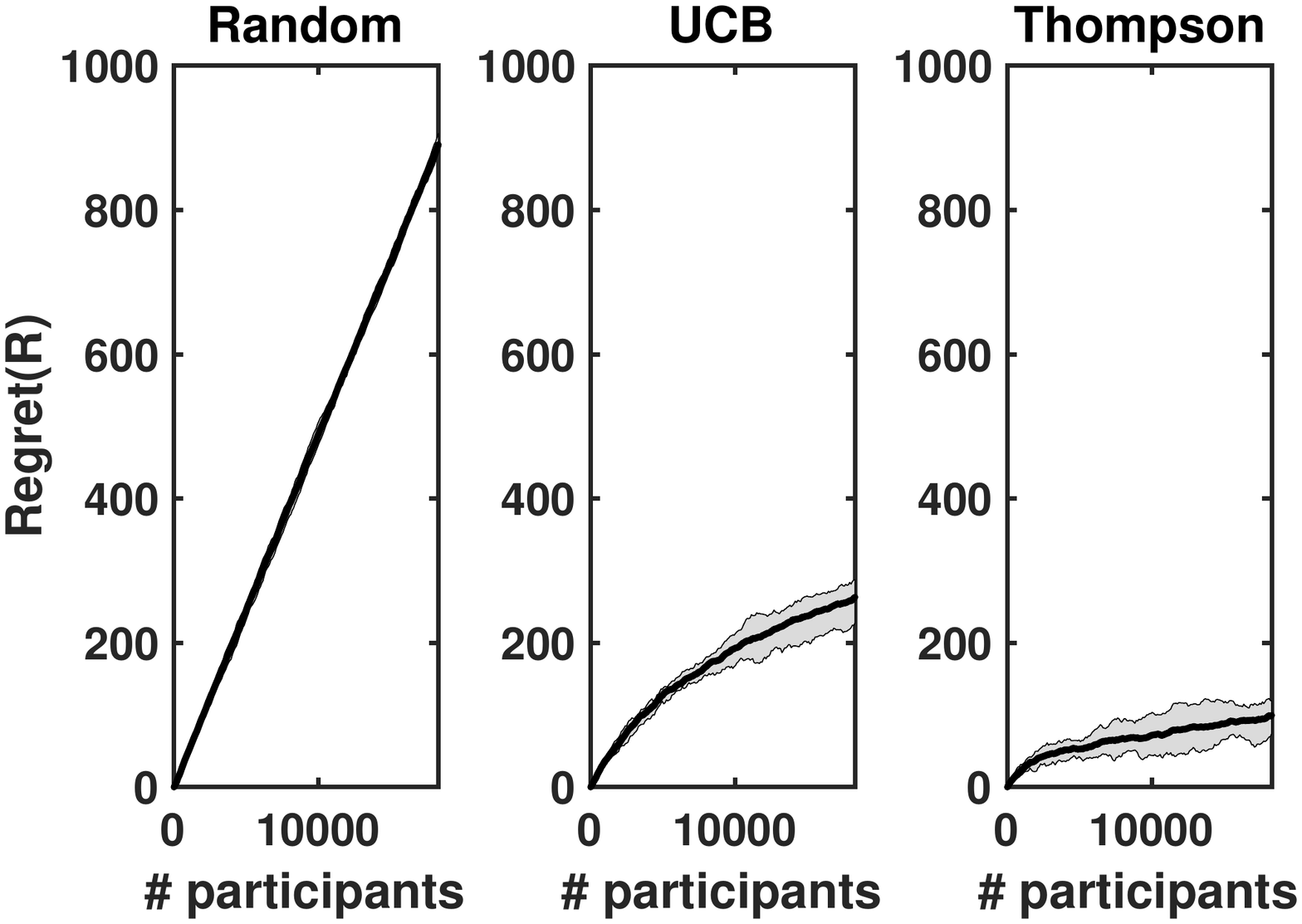}
		\caption{Regret: context-free multi-arm bandit.}
		\label{fig:result_mba_r}
	\end{subfigure}
	\begin{subfigure}[t]{0.49\textwidth}
		\centering
		\includegraphics[trim=0cm 6cm 0cm 6cm, clip, width=\linewidth]{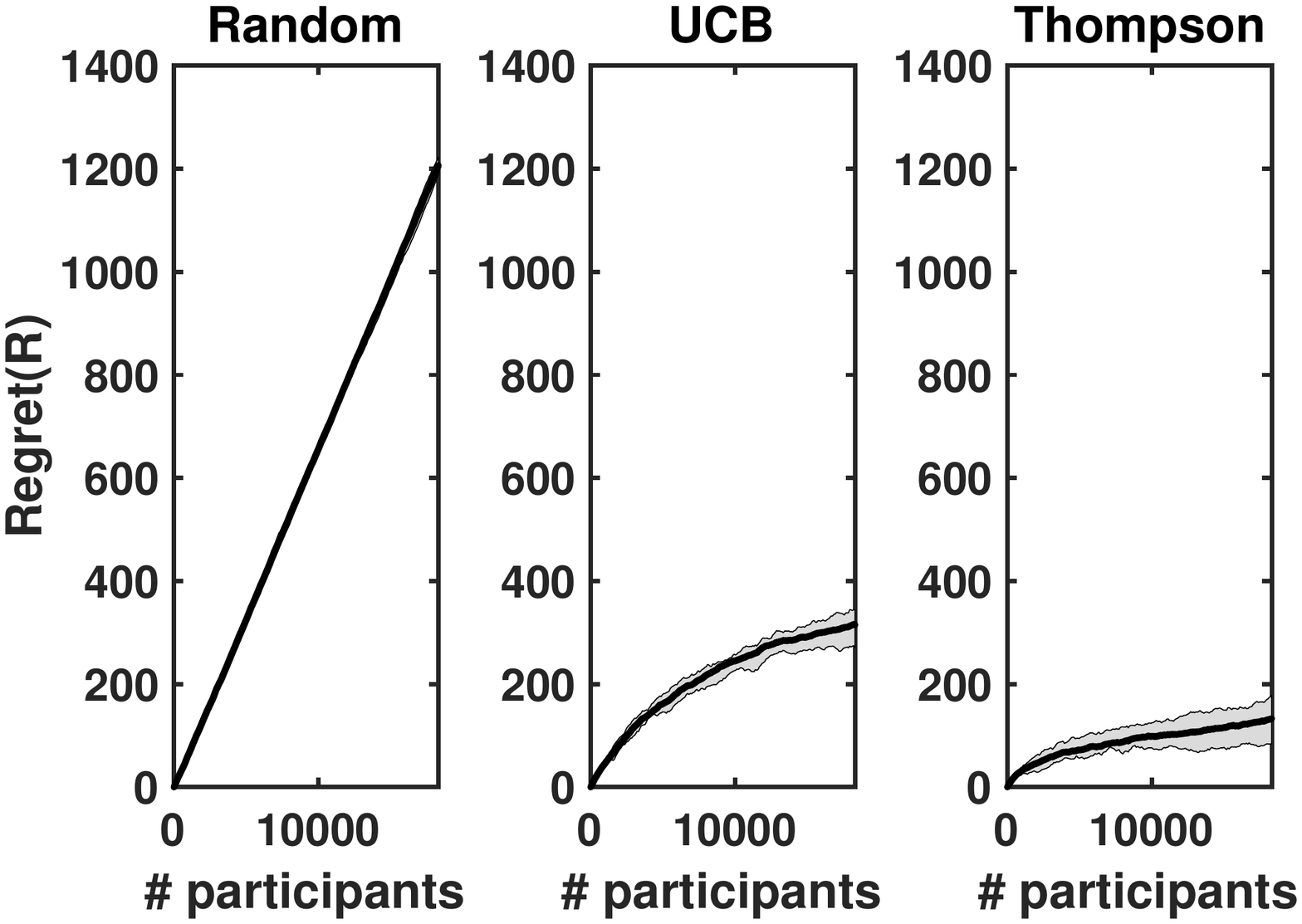}
		\caption{Regret: contextual bandit.}
		\label{fig:result_cb_r}
	\end{subfigure}
	\begin{subfigure}[t]{0.49\textwidth}
		\centering
		\includegraphics[trim=0cm 6cm 0cm 6cm, clip, width=\linewidth]{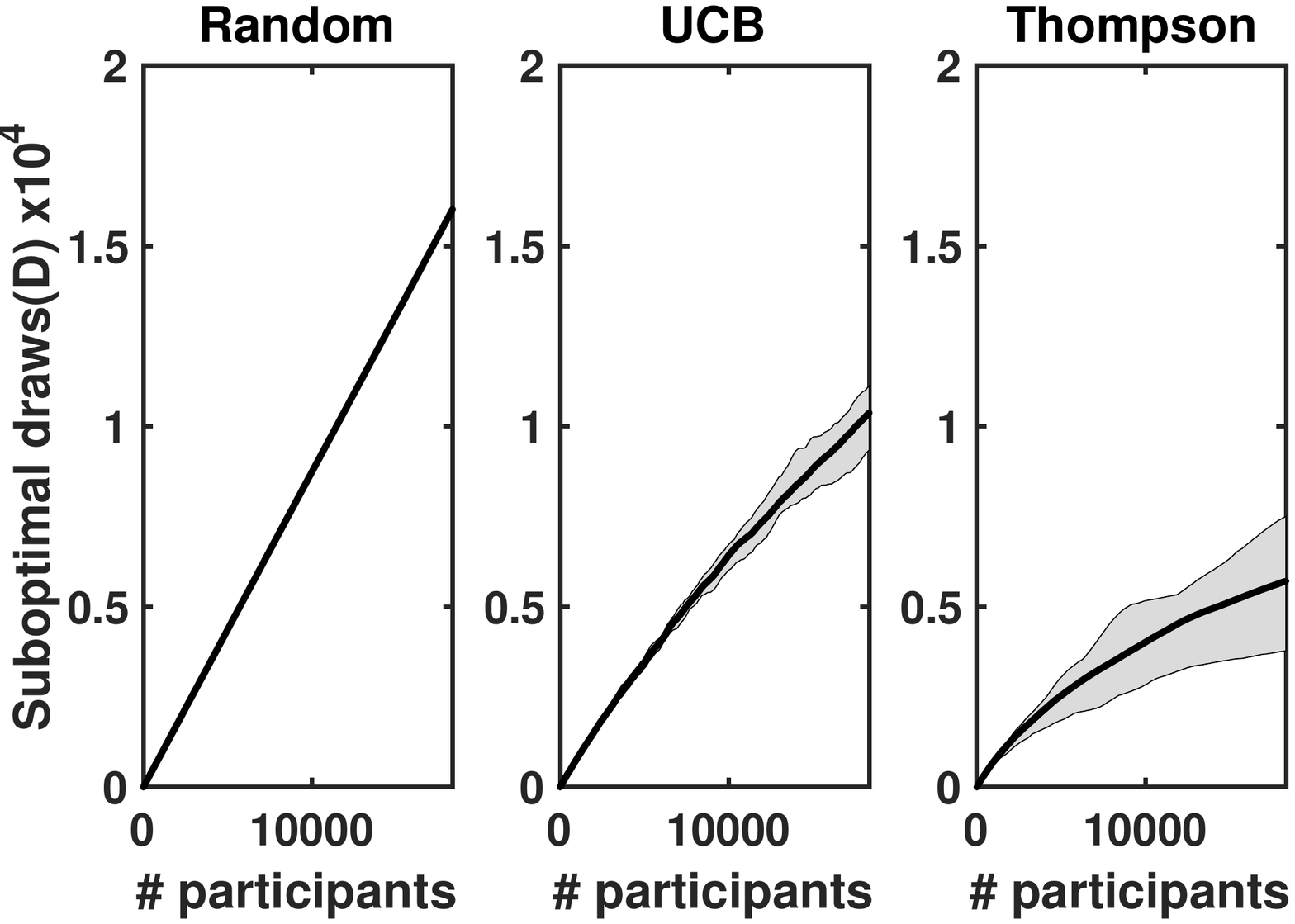}
		\caption{Suboptimal draws: context-free multi-arm bandit.}
		\label{fig:result_mba_d}
	\end{subfigure}
	\begin{subfigure}[t]{0.49\textwidth}
		\centering
		\includegraphics[trim=0cm 6cm 0cm 6cm, clip, width=\linewidth]{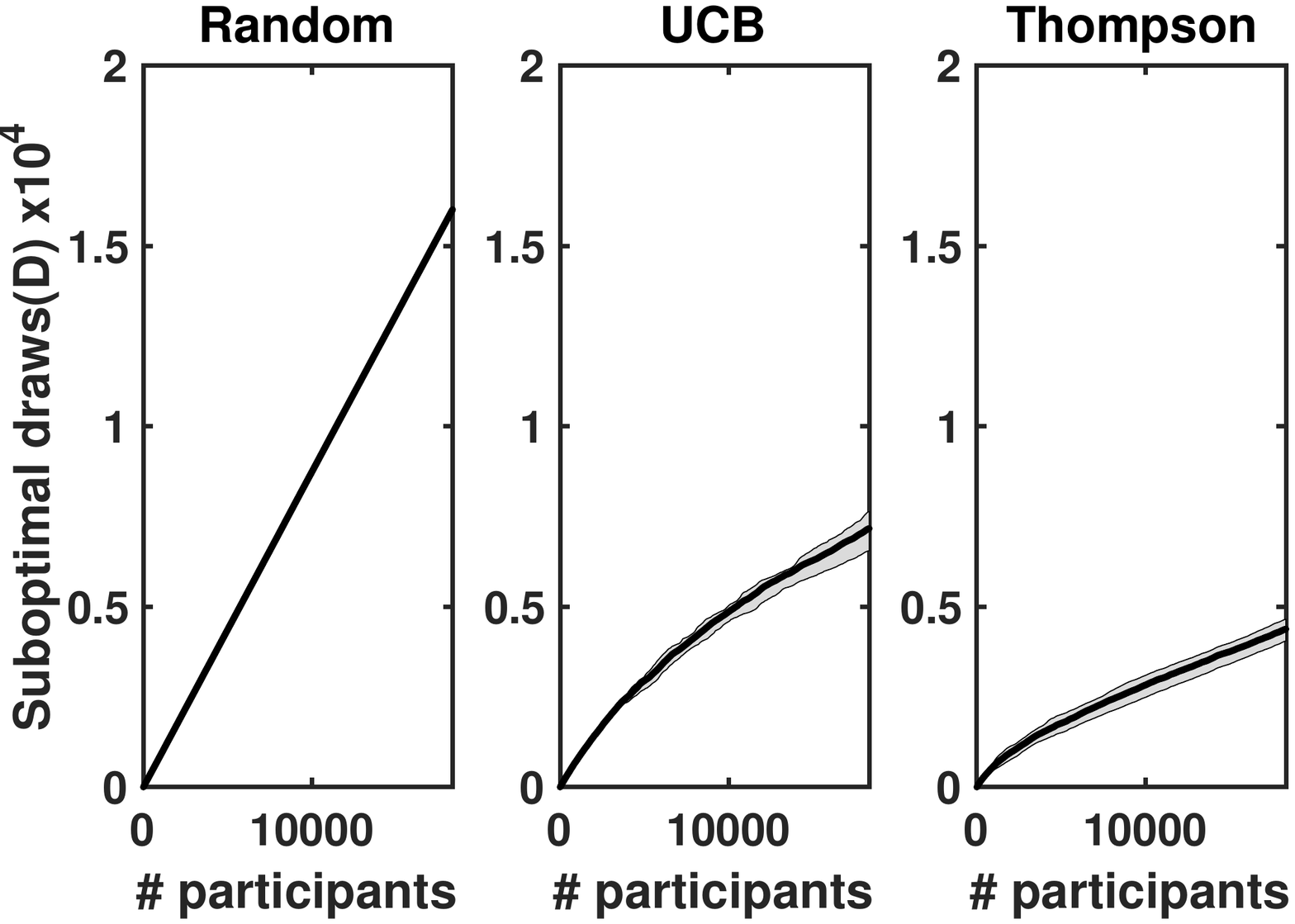}
		\caption{Suboptimal draws: contextual bandit.}
		\label{fig:result_cb_d}
	\end{subfigure}
	\caption{Regrets and suboptimal draw counts obtained using context-free multi-arm-bandit and contextual-bandit approaches for the IST database. Treatments were selected using random assignment, the UCB approach, and Thompson sampling. Plots include mean values of regrets and suboptimal draw counts and 25th percentile confidence intervals.}
	\label{fig:results_all}
\end{figure*}

Here we report the experimental results obtained using our contextual-bandit-based approach and a context-free multi-arm-bandit-based approach for the International Stroke Trial database. We ran each approach 20 times to evaluate the variability of regrets and suboptimal draw counts. Figure \ref{fig:results_all} shows the trend of cumulative regrets and suboptimal draw counts with increasing number of participants in the trial. Figures \ref{fig:result_mba_r} and \ref{fig:result_cb_r} show the cumulative regrets incurred in both the approaches when treatments were selected using random assignment, the UCB approach, and Thompson sampling. The plots include the mean regret values and the 25th percentile confidence intervals. Similarly, Figures \ref{fig:result_mba_d} and \ref{fig:result_cb_d} show the number of suboptimal draws in each case, as previously described. Table \ref{tab:results} reports the relative advantages of using the UCB approach and the Thompson sampling approach instead of random assignment, to select new treatments in the context-free multi-arm bandit and contextual-bandit cases. The relative advantages are illustrated as percentages of the regrets and suboptimal draw counts that were incurred in each case, compared with the random case. Note that the absolute regret values seen in Figures \ref{fig:result_mba_r} and \ref{fig:result_cb_r} are not comparable between the context-free and contextual approaches because they were calculated using different ground-truths (context-free and contextual ground-truths). However, the relative percentage improvements achieved using Thompson sampling and UCB approach, compared with the random approach, are comparable (since they were normalized using the respective random approaches).

\begin{table}[h]
	\centering
	\caption{The relative advantages of using the UCB approach and Thompson sampling to select new treatments, as opposed to using a random assignment, in the context-free multi-arm-bandit and contextual-bandit cases. For instance, the first entry in the table means that Thompson sampling incurs only 11.18$\pm$5\% of the regret incurred using a random assignment.}
	\label{tab:results}
	\begin{tabular}{@{}l|cccc@{}}
		\toprule\toprule
		\multirow{2}{*}{} & \multicolumn{2}{c}{Multi-arm bandit} & \multicolumn{2}{c}{Contextual bandit} \\ \cmidrule(l){2-5} 
		& Thompson          & UCB              & Thompson           & UCB              \\ \midrule
		Regret            & 11.18$\pm$5\%           & 29.57$\pm$7\%          & 11.03$\pm$3\%            & 26.10$\pm$4\%          \\
		Suboptimal draws  & 35.66$\pm$10\%           & 64.79$\pm$13\%          & 27.37$\pm$2\%            & 44.78$\pm$3\%          \\ \bottomrule
	\end{tabular}
\end{table}

\noindent \textbf{Significance:} It appears from Figure \ref{fig:results_all} that both the UCB and Thompson sampling approaches perform significantly better than random assignment in both the contextual-bandit and context-free multi-arm-bandit cases. Furthermore, the Thompson sampling approach seems to perform considerably better than the UCB approach in all the cases. When the contextual and the context-free cases are considered, Table \ref{tab:results} shows that the contextual case provides significant gains in the suboptimal draw counts and marginal gains in the regret value. Overall, our results indicate that the contextual-bandit-based approach that incorporates patient characteristics into the algorithm used to choose treatments for new participants in the study performs better than a context-free approach, is able to learn the differential response to treatments depending on the context of the participant (in this case whether or not the participant had atrial fibrillation) relatively quickly and provide significant advantages in correctly choosing treatments.
\section{Discussion}
\label{sec:disc}
Many factors contribute to the difficulty of developing and testing new therapies, including the difficulty of obtaining patient consent, variability in the standard of care, inadequate patient recruitment rates, and delays between trial phases as drugs move from early dose-finding to efficacy trials. An adaptive design is a statistical tool for accelerating drug development. Recent U.S. Food and Drug Administration draft guidance defines an adaptive design as a ``prospectively planned opportunity for modification of one or more specified aspects of the study design" based on interim analysis of a study \citep{us2010guidance}. The term ``prospective" means that modification is planned before data are examined in an unblinded manner. Prior to that definition, development of adaptive designs has a long and varied history. The idea of adaptive randomization  was introduced in the 1930s \citep{thompson1933likelihood}, sample size recalculation in the 1940s \citep{stein1945two}, sequential dose finding in the 1950s \citep{robbins1951stochastic}, and play-the-winner strategies and group-sequential methods in the 1960s \citep{zelen1969play}. However, there are costs associated with the use of adaptive designs, and they are seldom made obvious in the literature.

\noindent \textbf{Limitations and future work:} Our work has several limitations. First, this is a retrospective study, so the outcomes for treatments other than the ones actually provided to participants are unknown. Therefore, a model to simulate those outcomes was necessary to showcase the utility of our approach. The validity and comprehensiveness of the simulation model that we used in this work are debatable, considering the many confounder variables that might exist in a real scenario. Furthermore, we illustrated the utility of the model using a known risk-factor (i.e., atrial fibrillation) that can modulate response to stroke treatments. That significantly simplified our analyses because it reduced the number of contexts considered in our approach to two. However, in a real clinical trial setting, a myriad of clinical and biomarker variables would typically be collected, and the context space that included all these variables might explode. In addition, the number of participants required to achieve statistically significant gains in the regret and suboptimal draw counts will grow exponentially with the number of contexts in the model. Those are significant limitations that need to be answered before this model can be translated to a real clinical trial setting. We plan to investigate these limitations using function approximation methods that might eliminate the need for using as many bandits as the number of contexts in our model.
\section{Conclusion}
\label{sec:concl}
In this paper, we presented a \textit{contextual-bandit}-based online algorithm for optimizing treatment assignments in clinical trials. Unlike prior approaches, our algorithm takes patient characteristics into account in addition to the maximization of the clinical outcomes as criteria to determine treatments for new participants in the study. We evaluated our algorithm using a real clinical trial dataset from the International Stroke Trial. The results of our retrospective analysis indicate that the contextual-bandit-based approach performs significantly better than either a random assignment of treatments (the current gold standard) or a context-free multi-arm-bandit-based approach, providing substantial gains in the percentage of participants who receive the most suitable treatments. Hence, our study establishes the feasibility of an adaptive clinical trial setting that takes into account patient characteristics in adapting the trial attributes. The retrospective nature of the study and the difficulties of extending the model to the large number of clinical and biomarker variables collected in a clinical trial are some of the limitations of our approach, and future efforts will  address these limitations. 

{\small
	\bibliography{bibl}}

\end{document}